\documentclass[10pt,a4paper]{article}
\usepackage{iftex}
\ifPDFTeX
  \usepackage[T1]{fontenc}
  \usepackage{lmodern}
\else
  \usepackage{fontspec}
\fi
\usepackage[margin=1in]{geometry}
\usepackage{booktabs,array,longtable,amsmath,amssymb,enumitem,graphicx,xcolor,microtype,hyperref,url,caption,float,fancyhdr}
\hypersetup{colorlinks=true,linkcolor=blue!50!black,citecolor=blue!50!black,urlcolor=blue!50!black,pdfauthor={Shengzhou Kong},pdftitle={Omni-JEV: Text and Dual-Stream Speech Decisions}}
\setlist{nosep,leftmargin=*}
\title{\textbf{Omni-JEV: Text and Dual-Stream Speech Decisions}}
\author{%
  Shengzhou Kong\thanks{Corresponding author. Email: \href{mailto:kongsz@getui.com}{\nolinkurl{kongsz@getui.com}}.}\\[0.4em]
  \small Zhejiang Merit Interactive Research Institute Co., Ltd.
}
\date{}
\begin{document}
\maketitle
\begin{abstract}
A user utterance does not uniquely determine the next action in full-duplex dialogue: its consequences depend on the assistant's response state and interaction history. Speech activity, utterance completion, and interruption intent provide related evidence, whose implications for a response require an explicit action policy. We formulate this task at collected-segment endpoints and introduce Omni-JEV, which maps text or role-indexed user and assistant-reference speech to probabilities over five interaction actions. An aligned speech-to-speech understanding backbone encodes the acoustic roles separately and shares its candidate readout with text inputs. Bilingual weak supervision supports cross-entropy fine-tuning followed by reinforcement learning for probability-sensitive decisions (RLCD). We compare Gaussian-logit and categorical Choice RLCD from the same supervised initialization under matched training exposure, observing different trade-offs between classification and probability quality. Gaussian Omni-JEV achieves test accuracies of 84.08\% for audio and 96.55\% for text. This common action interface supports decisions about when a full-duplex voice assistant should respond, preserve ongoing speech, interrupt, or resume an interrupted response.
\end{abstract}
\textbf{Keywords:} Omni-JEV; interaction actions; speech-to-speech; dual-stream audio; action probabilities
\section{Introduction}
Full-duplex speech models let assistants listen during a response. Moshi and SALM-Duplex model both participants within speech generation~\cite{moshi2024,salmduplex2025}, and Freeze-Omni and MiniCPM-o develop integrated spoken interaction~\cite{freezeomni2024,minicpmo2026}. Incoming speech may acknowledge the current response, request a correction, or introduce a new request. Its consequences depend on both the contribution and the response it affects.

Research on turn taking provides increasingly informative predictions. Voice Activity Projection predicts future speech activity~\cite{vap2022}; FastTurn combines streaming semantics with acoustic cues for turn detection~\cite{fastturn2026}; and X2-Turn jointly predicts transcription and fine-grained user turn states~\cite{x2turn2026}. SID-Bench and its semantic interruption model distinguish substantive interruptions from backchannels~\cite{sidbench2026}. Such predictions already inform response control. Direct decision models offer a complementary interface: AudioJev returns candidate probabilities and studies next-speaker behavior from audio preceding proposed pauses~\cite{audiojev2026}, while Qwen-Audio-Jev supports typed choice, judgment, and scoring~\cite{qwenaudiojev2026}. These precedents incorporate semantics, local audio history, or typed probability decisions.

These prediction targets describe different aspects of the interaction. Utterance completion concerns the user's contribution, next-speaker prediction concerns subsequent behavior, and interruption detection concerns an active response. Relating that evidence to a response requires the assistant's state and history: a completed request can call for a new response while the assistant is idle or for stopping one already under way; resumption additionally depends on an earlier interruption. We study response actions defined relative to the assistant's state, using evidence available at a collected-segment endpoint. The research question is: \emph{how can text and role-indexed speech share this state-conditioned action target, and how does the quality of the learned distribution vary across optimization strategies and data sources?}

Ordinary automatic speech recognition (ASR) text does not fully preserve intonation, stress, duration, or overlap timing; FastTurn's acoustic--semantic modeling illustrates why lexical content alone may be insufficient for turn judgments~\cite{fastturn2026}. Keeping user and assistant-reference speech in separate role spans makes both the incoming contribution and its response context available to the model. A pause-confirmed or native segment endpoint fixes when evidence is available, while the target specifies how that evidence should affect the response.

We introduce \textbf{Omni-JEV}, which predicts \texttt{listen}, \texttt{start}, \texttt{continue}, \texttt{stop}, and \texttt{recover} within one action space. The model transfers the aligned acoustic frontend and Thinker from our separately trained speech-to-speech (STS) checkpoint, built on the MiniMind-O framework~\cite{minimindo2026}. Shared acoustic parameters encode user and assistant-reference waveforms into distinct role spans; text uses the same Thinker and candidate readout. A bilingual weak-label corpus from nine upstream source families connects observation endpoints, role evidence, response state, and action targets. Source-linked partitioning keeps related recordings and bilingual counterparts together.

A probability interface also requires evaluating more than the selected action. After cross-entropy supervised fine-tuning, we compare two continuations using reinforcement learning for probability-sensitive decisions (RLCD): Gaussian-logit exploration adapted from prior implementations~\cite{layasource,rsisource} and Qwen-Audio-Jev's categorical Choice strategy~\cite{qwenaudiojev2026}. They share initialization and training exposure. Gaussian achieves 84.08\% audio accuracy and 96.55\% text accuracy. Its audio macro-F1 is 76.00\%, versus 75.23\% for categorical, whereas categorical has higher text macro-F1 and lower probability errors in both modalities. This comparison characterizes the complete strategies and their classification--probability trade-offs.

Our contributions are:
\begin{enumerate}
 \item \textbf{A state-conditioned response-action task.} We specify how collected-segment evidence, assistant state, and history define five actions, and construct bilingual weak targets while distinguishing source annotations from derived action policies.
 \item \textbf{A shared model for text and role-indexed speech.} Omni-JEV adapts an aligned STS understanding path, separately represents user and assistant-reference audio, and uses one candidate-probability readout for both modalities.
 \item \textbf{An empirical characterization of action distributions.} Matched Gaussian and categorical RLCD continuations reveal different classification--probability trade-offs, while source-level evaluation identifies the scope of the aggregate results.
\end{enumerate}

\section{Related Work}\label{sec:related}
\subsection{From turn events to action decisions}
The distinction between detecting an event and choosing a response is central to interaction control. Voice Activity Projection predicts joint voice activity, while FastTurn combines acoustic and streaming semantic cues for turn decisions~\cite{vap2022,fastturn2026}. AudioJev instead exposes candidate probabilities and studies their consistency under alternative orderings~\cite{audiojev2026}. Qwen-Audio-Jev provides Choice, Noul, and Score interfaces, including categorical RLCD for decision distributions~\cite{qwenaudiojev2026}. Omni-JEV adopts candidate-based prediction for state-conditioned interaction actions at collected-segment endpoints. Its candidate permutations vary symbol--action correspondences during training; empirical order invariance remains a separate property.

\subsection{Speech understanding and role representations}
Two representation choices are relevant to response actions: how speech reaches a language model, and how participant roles remain distinguishable. MiniMind-O provides our Thinker--Talker framework~\cite{minimindo2026}, with SenseVoice and Qwen3 supplying the acoustic and language components~\cite{funaudiollm2024,sensevoicesource,qwen3report2025}. Qwen3-Omni, Step-Audio 2, MOSS-Audio, and FireRedAudio provide broader audio-language backbones~\cite{qwen3omnireport2025,stepaudio22025,mossaudio2026,fireredaudio2026}. Freeze-Omni connects speech modules to a language model; Moshi, SALM-Duplex, MiniCPM-o 4.5, and DuplexSLA address parallel streams, streaming interaction, or structured actions~\cite{freezeomni2024,moshi2024,salmduplex2025,minicpmo2026,duplexsla2026}. Channel-routing work explicitly examines user and assistant streams~\cite{routingduplex2026}. Omni-JEV uses a shared encoder but separate role spans: its output is an action distribution at an observed endpoint, rather than the next generated speech segment.

Full-Duplex-Bench and its multi-turn successor evaluate behavior across time, and HumDial-FDBench provides human-recorded dual-channel dialogues~\cite{fdbench2025,fdbenchv2_2026,humdial2026}. Our endpoint benchmark measures one component of this broader problem. Its intent, command, environmental, and conversational sources~\cite{banking77,massive,minds14,speechcommands,escsource,wakeupsource,fdv2source,fdv3source,smoothconvsource} supply heterogeneous evidence; Section~\ref{sec:data} distinguishes their original annotations from derived action targets.

\subsection{Learning action distributions}
Gaussian RLCD and categorical Choice differ in what is explored and scored: Gaussian perturbs relative logits and evaluates a complete probability vector, whereas Choice samples candidate positions to estimate negative Brier~\cite{qwenaudiojev2026}. The underlying estimator and scoring principles come from prior work~\cite{reinforce1992,proper2007,layasource,rsisource}; Appendix~\ref{app:rlcd-details} records the inherited mechanisms and recipe choices. Our comparison applies both complete strategies to the same Omni-JEV parent. This differs from optimizing generated speech trajectories with interaction and semantic rewards, as in Multi-Faceted Interactivity Alignment~\cite{interactivity2026}: here labeled distributions supply rewards after one model forward pass.

\section{Interaction Decision Task}\label{sec:task}
\subsection{State-conditioned action prediction}\label{sec:decision-problem}
Let $e$ be the endpoint of a collected user segment, $s_e$ the interaction state, and $h_e$ the available history. The observation is text context or a pair of aligned acoustic prefixes:
\begin{equation}
 x_e^{\mathrm{text}}=(s_e,h_e,a_e,u_e),\qquad
 x_e^{\mathrm{audio}}=(s_e,h_e,U_{b:e},A_{b:e},m),
\end{equation}
Here $a_e$ and $u_e$ are assistant and user text, $U_{b:e}$ and $A_{b:e}$ are waveforms over a common interval, and $m$ records channel availability. The model predicts $p_\theta(c\mid x_e)$ over the presented candidates $C$. Dual-stream audio denotes role-indexed inputs: the assistant track is a reference of its response, and the user track supplies the new interaction evidence.

For conversational sources, $s_e$ is reconstructed from observations through $e$, before applying the current target action; it is not device playback telemetry. Short-form sources instead supply a task-given or added scenario, or an unknown state. Table~\ref{tab:input-context} specifies the fields actually serialized by the model. History $h_e$ is carried by the role prefixes, retained state flags, and any completed text history before the audio window; it is not an additional inferred transcript.

\begin{table}[htbp]
\centering\small
\caption{Context supplied to the action model. Optional fields retain available source information; an absent field does not assert a negative state.}
\label{tab:input-context}
\begin{tabular}{@{}p{0.25\linewidth}p{0.69\linewidth}@{}}\toprule
Field & Content and interpretation\\\midrule
\texttt{text} & User/assistant text for text examples; optional completed history preceding the audio window for speech examples. Current-window ASR text is excluded.\\
\texttt{assistant\_state} & Source state in \texttt{value}, such as \texttt{idle}, \texttt{playing}, \texttt{interrupted}, or \texttt{unknown}; optional flags retain pending interruption, resumability, and assistant speech during the collected segment.\\
\texttt{policy}, \texttt{scenario} & Available policy text and explicit scenario conditions, including addressing or resumability assumptions. These may be empty.\\
Role availability & Supplied user and assistant-reference spans; a missing role is stated in the prompt and its acoustic span is omitted.\\\bottomrule
\end{tabular}
\end{table}

The target describes how the assistant should act on the collected segment. It is distinct from a domain intent: a banking query can require starting a response or stopping an existing one, depending on state. Table~\ref{tab:actions} defines the common five-action policy. In particular, \texttt{continue} preserves an active response, whereas \texttt{listen} defers a response when none is to be preserved or resumed. The source-specific construction rules, including isolated-word exceptions, appear in Appendix~\ref{app:annotation}.

\begin{table}[htbp]
\centering\small
\caption{Interaction actions associated with a collected user segment. The response active during the segment determines whether an interruption is applicable.}
\label{tab:actions}
\begin{tabular}{l p{0.76\linewidth}}\toprule
Action & Operational interpretation\\\midrule
\texttt{listen} & Await more evidence when no response is to be preserved or resumed.\\
\texttt{start} & Begin a response to the user when no existing response should be preserved.\\
\texttt{continue} & Preserve a response that remains active; the segment does not warrant interruption.\\
\texttt{stop} & Interrupt a response active during the segment because of the user's intent.\\
\texttt{recover} & Resume a response after a preceding stop or recoverable interruption.\\\bottomrule
\end{tabular}
\end{table}

\subsection{Causal prefixes and decision endpoints}
Speech onset starts collection; a pause-confirmed VAD endpoint triggers the decision. Reliable native segment ends also supply construction endpoints. The common interval $[b,e)$ retains the role history used to interpret the collected segment. Its anchor follows the source annotation: observed response turns for Wakeup and FD, and the preceding rule-labeled \texttt{start} segment for SmoothConv. Before a usable anchor exists, history begins at the recording start. Appendix~\ref{app:annotation} specifies these offline construction policies.

This observation boundary separates annotation from inference. Later recording events may help annotate what the collected segment meant, but they are excluded from its model input. A \texttt{stop} target concerns a response active during the segment; the endpoint state can already be \texttt{idle} if that response ended before the collected user segment ends. Available interval-speech flags distinguish this fact from endpoint playback, and dispatch occurs after collection. Offline experiments use complete corpus prefixes. The online implementation additionally truncates older history at completed boundaries; its policy is specified in Appendix~\ref{app:evidence}.

\begin{figure}[t]
\centering
\includegraphics[width=\linewidth]{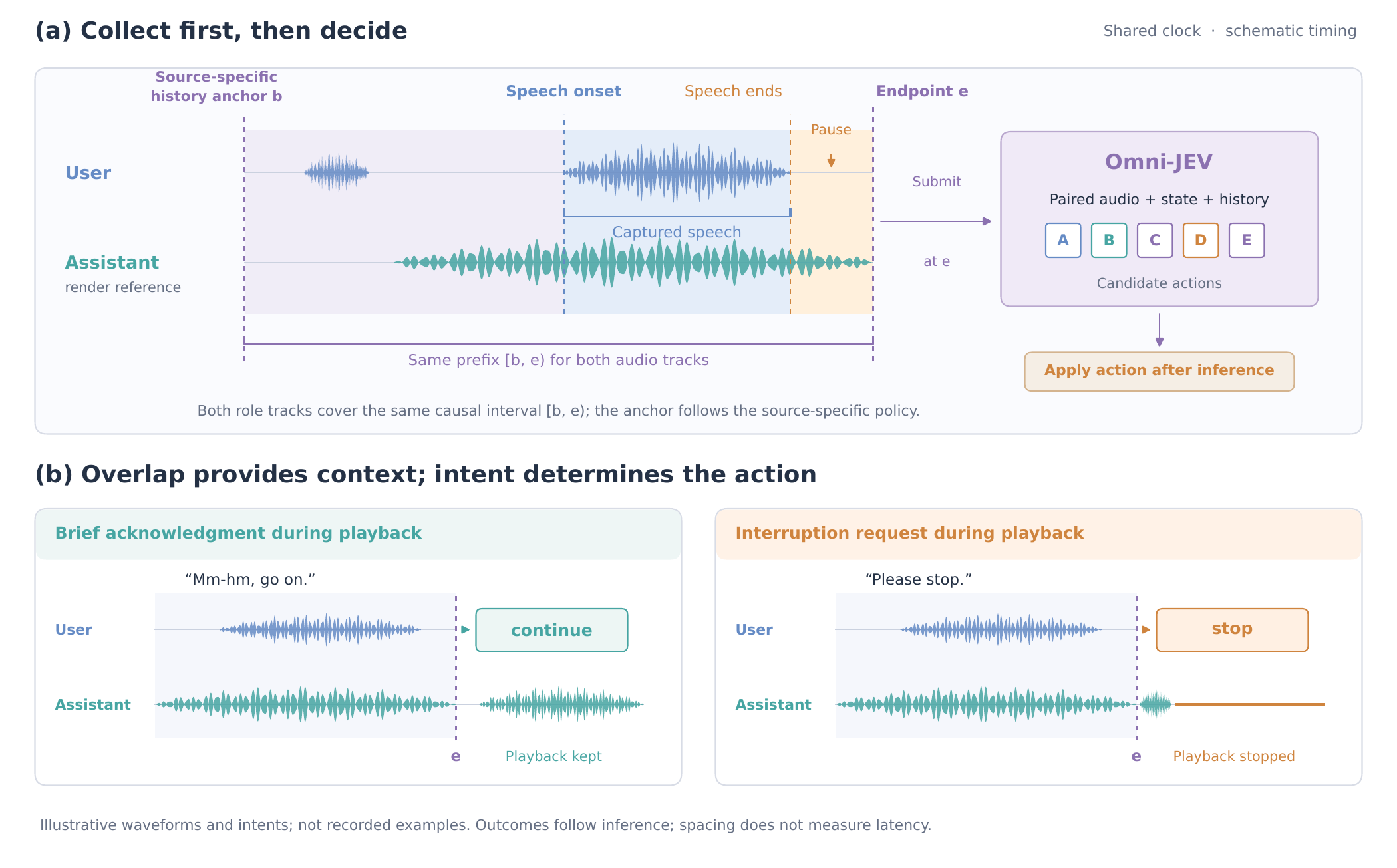}
\caption{Interaction decisions at collected-segment endpoints. Acknowledgment and interruption can both overlap assistant speech but require different actions. Both role tracks share the source-specific history anchor and contain only evidence available through the endpoint; subsequent behavior illustrates the action semantics.}
\label{fig:vad-actions}
\end{figure}

\section{Omni-JEV Architecture}\label{sec:model}
\begin{figure}[t]
\centering
\includegraphics[width=\linewidth]{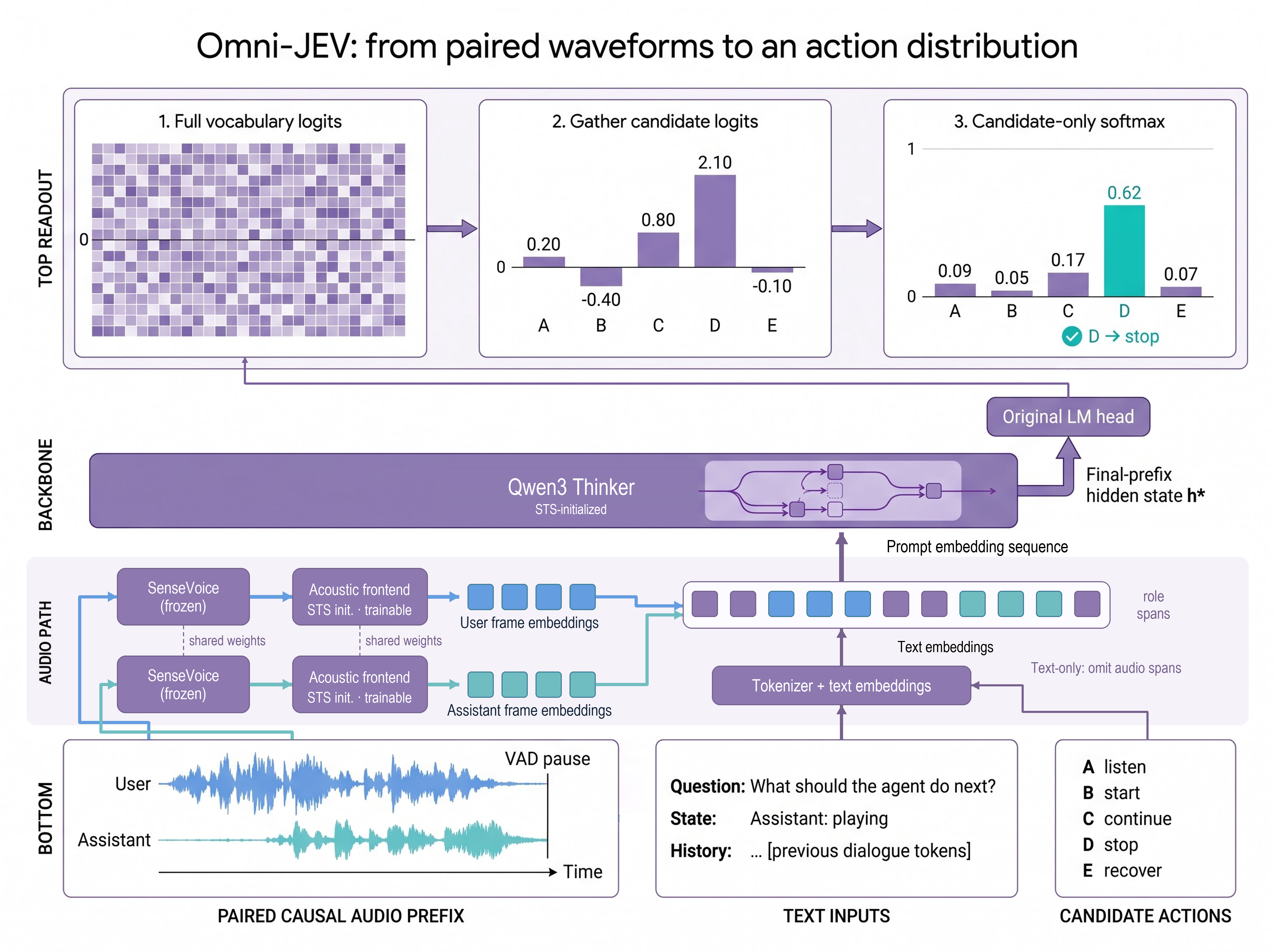}
\caption{Omni-JEV architecture. User and assistant-reference audio are independently encoded with shared SenseVoice and frontend parameters, then serialized with text into one prompt. The STS-initialized Thinker and original LM head produce a distribution over candidate actions. Colors distinguish roles and text; A--E and the readout values show one example ordering.}
\label{fig:architecture}
\end{figure}

Omni-JEV implements the task in Section~\ref{sec:task} through a shared understanding path and action readout. Figure~\ref{fig:architecture} connects its three components: an STS-initialized acoustic frontend and Thinker, a sequence that retains user and assistant roles, and a candidate distribution from the original vocabulary head. Both text and speech produce a decision at the same answer position; text-only inputs omit the acoustic spans.

\subsection{An aligned STS understanding path}\label{sec:sts-foundation}
Our STS implementation builds on the MiniMind-O Thinker--Talker framework~\cite{minimindo2026}, connecting a SenseVoice encoder~\cite{funaudiollm2024,sensevoicesource}, a learned acoustic frontend, a Qwen3-based Thinker~\cite{qwen3report2025}, and a speech-generation Talker. Omni-JEV reuses the frontend and Thinker after acoustic-to-language alignment. The transferred frontend maps speech representations into the language-model embedding space; action supervision subsequently adapts that mapping and the Thinker to the interaction policy.

The frontend and Thinker contain 597,625,856 trainable parameters, with hidden width $d=1024$; SenseVoice is additional and frozen during action adaptation. Appendix~\ref{app:initialization} specifies the checkpoint, provenance, and full configuration. The Talker is absent from the decision computation: the adapted understanding path selects an action without first generating speech.

\subsection{Role-preserving acoustic representation}
The action depends on the relationship between incoming speech and the assistant's response. To preserve that distinction, we encode each available role independently with the same encoder and frontend. Each waveform is resampled to 16\,kHz. For role $c\in\{U,A\}$, let $H_c\in\mathbb R^{T_c\times512}$ contain its valid encoder frames. The frontend maps these frames to the Thinker's hidden width $d$:
\begin{equation}\label{eq:frontend}
 H_c=E_{\mathrm{SV}}(X_c),\qquad
 V_c=\operatorname{GELU}\!\left(\operatorname{LN}(H_c)W_1+b_1\right)W_2+b_2,
\end{equation}
The projection matrices have shapes $W_1\in\mathbb R^{512\times d}$ and $W_2\in\mathbb R^{d\times d}$. Shared weights place both roles in the same acoustic representation space, while separate spans retain their identities. The waveforms cover the same temporal interval, preserving the reference window for silence and speech. The final decision state can attend to both encoded roles within the shared Transformer. The quality of that distinction depends on the source channels, as discussed in Section~\ref{sec:data}.

\subsection{One sequence for text and speech}
The prompt serializes the instruction, role-indexed audio, interaction context, candidate descriptions, and answer prefix. Each audio span contains one placeholder per valid frame; its embedding is replaced by the corresponding row of $V_U$ or $V_A$:
\begin{equation}\label{eq:sequence}
 \mathcal E_e=\operatorname{concat}
 \bigl(E_{\mathrm{instr}},E_{\mathrm{role}:U},V_U,
 E_{\mathrm{role}:A},V_A,E_{\mathrm{ctx}},E_C,E_{\mathrm{ans}}\bigr).
\end{equation}
These positions are continuous frame embeddings. The instruction and candidate descriptions define the requested decision, the context serializes the fields in Table~\ref{tab:input-context}, and the acoustic spans supply the observed speech. Text and audio enter the same causal Transformer, without a separate channel-fusion block. When audio is unavailable, the prompt states the missing role explicitly and omits its span. Text-only examples omit both acoustic spans while retaining context and candidate fields. This construction gives the two modalities the same prediction target and readout.

\subsection{Action probabilities from the original LM head}
Let $h_e^*$ be the hidden state at the final answer-prefix position. The original vocabulary head produces $\ell_e=W_{\mathrm{LM}}h_e^*$. For candidate $C_j$ assigned token ID $v_j$, we gather and normalize only the candidate logits:
\begin{equation}\label{eq:readout}
 z_j=\ell_e[v_j],\qquad
 p_\theta(C_j\mid x_e)=\frac{\exp z_j}{\sum_{k=1}^{|C|}\exp z_k}.
\end{equation}
The symbols A--E denote candidate positions, not fixed actions. During training, candidates and target positions are permuted together. Each symbol is a single token at the answer boundary, allowing direct readout without generating an explanation or adding a new classifier. Both modalities backpropagate through this head and the shared Thinker; audio examples also update the frontend. Inference returns the unperturbed candidate softmax and its action mapping to the controller.

\section{Training Omni-JEV}\label{sec:training}
\begin{figure}[t]
\centering
\includegraphics[width=\linewidth]{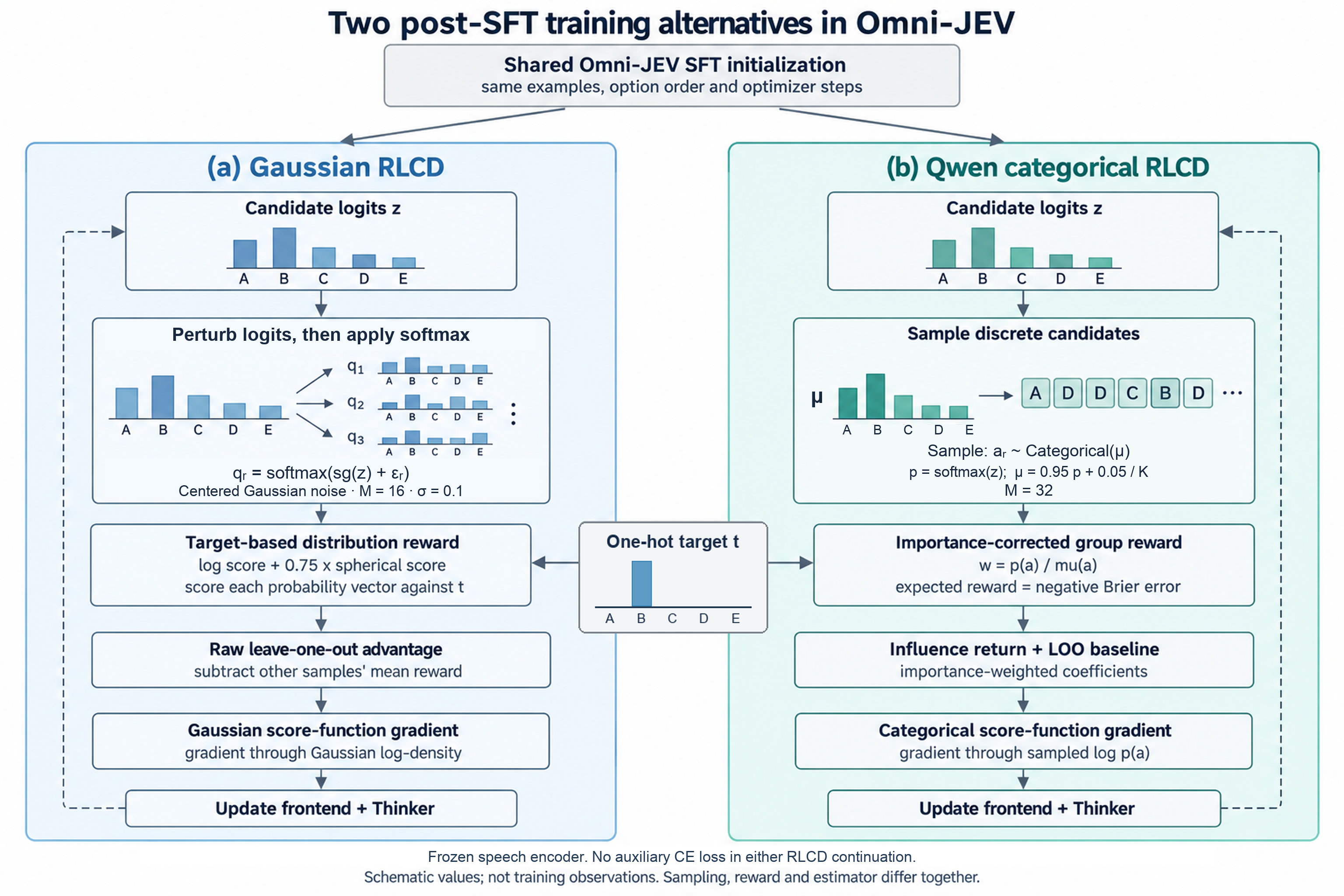}
\caption{RLCD continuations from one Omni-JEV SFT checkpoint. Gaussian scores perturbed probability vectors; categorical Choice samples candidate positions to estimate negative Brier~\cite{qwenaudiojev2026}. Labels supply rewards; bars and samples are schematic. The columns compare complete strategies.}
\label{fig:rlcd-recipes}
\end{figure}

Cross-entropy supervision adapts the STS understanding path to the action task; RLCD then optimizes candidate distributions against the same targets. Figure~\ref{fig:rlcd-recipes} contrasts Gaussian exploration of complete probability vectors with categorical sampling of candidate positions. Evaluation uses the unperturbed distributions.

\subsection{Supervised action adaptation}
For candidate logits $z$ and target $t\in\Delta^{K-1}$, SFT minimizes
\begin{equation}
 \mathcal L_{\mathrm{SFT}}=-\sum_{k=1}^{K}t_k\log p_\theta(k\mid x).
\end{equation}
Targets are one-hot over $K=5$ positions and follow each example's candidate permutation. One supervised epoch establishes the common frontend and Thinker initialization for both continuations.

\subsection{Gaussian exploration of action distributions}
A perturbation explores relative candidate preferences, allowing rewards to evaluate a complete probability vector. With $P=I-\mathbf1\mathbf1^\top/K$ and independent $\xi_r\sim\mathcal N(0,I)$, define
\begin{equation}\label{eq:gaussian}
 \epsilon_r=\sigma P\xi_r,\qquad
 \widetilde z_r=\operatorname{sg}(z)+\epsilon_r,\qquad
 q_r=\operatorname{softmax}(\widetilde z_r),\quad r=1,\ldots,M.
\end{equation}
Here $\operatorname{sg}$ stops gradients. Centering removes the common-shift direction to which softmax is invariant, without changing the sampled probability distribution. We score target agreement using the inherited log-plus-spherical reward~\cite{layasource,proper2007}:
\begin{equation}\label{eq:reward}
 R(q_r,t)=\sum_k t_k\log q_{rk}
       +\beta\frac{\langle t,q_r\rangle}{\|q_r\|_2},\qquad \beta=0.75.
\end{equation}
The log term rewards target probability; the spherical term measures target alignment relative to the vector norm. Each sample remains a normalized distribution over all actions, so the reward evaluates both target mass and its relation to competing probabilities. The reward uses labeled targets during training; it is not a reward obtained from a generated dialogue trajectory. The Gaussian exploration and density-gradient construction follow prior RLCD implementations~\cite{layasource,rsisource}; our adaptation uses centered noise and raw leave-one-out advantages. Appendix~\ref{app:rlcd-details} details this provenance and the reward's scoring properties.

\subsection{Score-function optimization}
Each perturbation is compared with the others for the same input:
\begin{equation}\label{eq:gaussian-advantage}
 A_r=R(q_r,t)-\frac{1}{M-1}\sum_{s\ne r}R(q_s,t),
\end{equation}
We retain the raw advantage scale. Treating sampled vectors, rewards, and advantages as constants gives the score-function update~\cite{reinforce1992}
\begin{equation}\label{eq:gaussian-gradient}
 \widehat g_z=\frac{1}{M\sigma^2}\sum_{r=1}^{M} A_r\epsilon_r,
 \qquad \nabla_\theta\mathcal L_G
 =-\left(\frac{\partial z}{\partial\theta}\right)^\top\widehat g_z.
\end{equation}
The gradient passes through the Gaussian density's center, rather than through sampled rewards. A perturbation with positive advantage contributes in its sampled direction, while a negative advantage contributes in the opposite direction. Their average updates the relative candidate logits, and the model Jacobian transfers that update to the shared frontend and Thinker. The update targets a locally smoothed reward; its effect on the unperturbed inference distribution is evaluated empirically. One encoder/Thinker forward pass supplies the logits; subsequent samples require only vector calculations. We use $M=16$ and $\sigma=0.1$, without auxiliary CE or reference-preservation losses. Appendix~\ref{app:rlcd-details} gives the surrogate loss, its Gaussian-smoothed objective, and the estimator's scope; numerical clamps appear in Appendix~\ref{app:evidence}.

\subsection{Categorical RLCD comparison}
We reproduce Qwen-Audio-Jev's Choice formulation~\cite{qwenaudiojev2026}, targeting negative Brier error:
\begin{equation}\label{eq:categorical-objective}
 J_C(p,t)=-\|p-t\|_2^2.
\end{equation}
With $p=\operatorname{softmax}(z)$, candidates are drawn from $\mu=(1-\eta)p+\eta/K$ and importance-weighted by $w_r=p(a_r)/\mu(a_r)$. A pairwise collision term estimates the squared probability norm; detached influence returns and a leave-one-out baseline weight sampled log-probability gradients. We use $M=32$, $\eta=0.05$, and no ordinal term ($\lambda=0$). The complete objective and update appear in Appendix~\ref{app:rlcd-details}.

Both continuations return the unperturbed candidate softmax at inference. Their rewards, exploration spaces, and estimators differ together, so the matched protocol in Section~\ref{sec:rlcd-protocol} compares complete strategies rather than isolating an estimator.

\section{A Bilingual Corpus for Interaction Actions}\label{sec:data}
An interaction target requires more than an utterance label: the same content can imply different actions under different response states. We therefore construct examples around a common unit comprising the observed segment, its context, role availability, and a five-action target.

\subsection{Sources and action construction}
Table~\ref{tab:dataset} summarizes nine upstream source families in ten displayed rows: FD v2 and FD v3 are shown separately because their provenance and license records differ. BANKING77 and MASSIVE provide intent text; Mini Speech Commands and MInDS-14 provide spoken commands and intents; ESC-10 provides environmental sounds~\cite{banking77,massive,speechcommands,minds14,escsource}. An inherited turn-text corpus supplies additional interaction contexts~\cite{legacytextdata}. The construction distinguishes observed state from an added labeling scenario. BANKING77 and MASSIVE utterances are presented to a text teacher under an assumed idle state with no resumable response; the teacher assigns an interaction action rather than copying the domain-intent ID. The inherited turn-text task supplies assistant/user text and a task-given playing state. Short-form audio uses explicit scenario rules: MInDS-14 requests yield \texttt{start} under an idle, completed-request assumption; ESC-10 clips yield \texttt{listen} without a sound-identification or alarm task. Mini Speech Commands instead uses a word-to-action convention with unknown assistant state. These constructed conditions differ from the observed turn state in conversational sources. Table~\ref{tab:construction} gives the source-to-action mapping and label-inheritance conditions.

Intelligent Wakeup, FD\_data\_v2/v3, and SmoothConv supply conversational evidence~\cite{wakeupsource,fdv2source,fdv3source,smoothconvsource}. SmoothConv provides native role/channel annotations; FD\_data\_v2 is described by its upstream card as a synthetic conversation source with synchronized role tracks, while FD\_data\_v3 is retained as a separate upstream source; Wakeup provides original speaker timing annotations and uses user VAD endpoints. Original annotations, deterministic state rules, and saved ASR/LLM-assisted decisions produce the action targets under the policy in Section~\ref{sec:task}. The resulting action targets are derived weak supervision. The sources do not share one annotation algorithm: SmoothConv uses native segment ends and ordered temporal rules, whereas the FD pause pipeline uses saved causal ASR and DeepSeek-nothink judgments. Wakeup retains original speaker identity and reuses existing semantic targets while aligning Chinese endpoints separately. Appendix~\ref{app:annotation} distinguishes these procedures and their remaining documentation limits.

\subsection{Licensing and release boundary}
The corpus combines sources with heterogeneous licensing and provenance conditions. BANKING77, MASSIVE, MInDS-14, and the ESC-10 subset are used with their respective attribution requirements; the local Speech Commands card points to the upstream dataset and its license information. The FD v2 card declares Apache-2.0, but the current archive does not retain an independent license record for FD v3, so the combined FD v2/v3 source is not described as uniformly Apache-2.0. Intelligent Wakeup and SmoothConv declare CC BY-NC 4.0; their use in this work is restricted to the noncommercial research setting. The inherited turn-text corpus is an internal, unpublished record rather than a separately licensed public benchmark, and its underlying text ownership and external publication permission are not established by the current ledger.

Translation, relabeling, boundary alignment, and TTS synthesis are treated as derived processing steps. They do not remove the licensing, privacy, or release conditions of the underlying sources. We report aggregate results and retain source-linked provenance, but do not distribute the raw recordings, derived bilingual audio, complete action-label corpus, or model weights with this manuscript. Any commercial training, deployment, or public release requires a separate review of the applicable upstream and internal permissions.

\begin{table}[htbp]
\centering\small
\caption{Composition of the bilingual interaction corpus. Language denotes the version used in this work; the none entry denotes nonlinguistic audio.}
\label{tab:dataset}
\begin{tabular}{lllrr}
\toprule
Integrated source & Modality & Language & Train & Test\\
\midrule
BANKING77~\cite{banking77} & text & en/zh & 22,726 & 2,509\\
MASSIVE~\cite{massive} & text & en/zh & 27,491 & 3,034\\
Legacy turn text~\cite{legacytextdata} & text & en/zh & 45,570 & 5,030\\
Mini Speech Commands~\cite{speechcommands} & audio & en/zh & 14,312 & 1,580\\
MInDS-14~\cite{minds14} & audio & en/zh & 904 & 100\\
ESC-10~\cite{escsource} & audio & none & 360 & 40\\
Wakeup~\cite{wakeupsource} & audio & en/zh & 66,687 & 7,380\\
FD v2~\cite{fdv2source} & audio & en/zh & 492 & 66\\
FD v3~\cite{fdv3source} & audio & en/zh & 1,158 & 118\\
SmoothConv~\cite{smoothconvsource} & audio & zh & 133,786 & 15,003\\
\textbf{Total} & -- & -- & \textbf{313,486} & \textbf{34,860}\\
\bottomrule
\end{tabular}

\end{table}

\subsection{Role availability and bilingual alignment}
We preserve native separated tracks when available and mark absent assistant-reference audio explicitly. For mono recordings, role timing masks retain role-indexed evidence but do not separate voices during overlap: both inputs may contain the same mixture. Thus dual-stream support describes the input interface, while channel provenance determines the strength of role evidence in each example.

Chinese and English counterparts retain source links. Translated speech is aligned to its own segment boundaries; it does not inherit source-language timestamps. Subsequent events may inform offline labeling, but the constructed prefix ends at the decision endpoint. ASR and translation support corpus construction without making transcripts mandatory for audio inference.

\subsection{Source-linked partition}
We group recording identities, bilingual counterparts, and exact normalized input links into 57,240 connected components. Seeded component allocation, jointly stratified by source, language, modality, and action, produces a two-way split with approximately 90\% training and 10\% testing (seed 20261023). All linked examples remain on one side.

The corpus contains 313,486 training and 34,860 test instances. Audio contributes 217,699 training and 24,287 test instances; text contributes 95,787 and 10,573. There is no validation split. Eligibility exclusions and action support are given in Appendix~\ref{app:support}. Preserving source groups reduces known cross-split links, while class imbalance and unknown semantic near-duplicates remain properties of the data.

\section{Experimental Setup}\label{sec:protocol}
We address three questions: \textbf{RQ1:} How does Omni-JEV compare with external systems on text and speech action recognition under the stated inputs? \textbf{RQ2:} How do matched Gaussian and categorical continuations differ in classification and probability quality? \textbf{RQ3:} How do source and language composition affect aggregate scores?

\subsection{Training and evaluation}
The source-linked split in Section~\ref{sec:data} supplies one natural-frequency SFT epoch and the fixed test set. Offline inference includes every test example with complete, untruncated audio prefixes under a 40,960-token context limit, independently of the online history-window policy. Final checkpoints follow the prescribed budget; both continuations use one training seed. Optimizer and numerical settings appear in Appendix~\ref{app:evidence}.

\subsection{Matched RLCD comparison on Omni-JEV}\label{sec:rlcd-protocol}
Both branches share the SFT checkpoint, ordered examples, candidate permutations, inputs, trainable parameters, and optimizer schedule. Each processes all 313,486 training examples once in 78,372 optimizer updates, with accumulation four and a final window of two examples. Neither adds auxiliary CE. Their final checkpoints are evaluated on the same 24,287 audio and 10,573 text examples. Holding the supervised parent fixed makes the comparison specific to continuation on Omni-JEV, rather than a comparison of separately pretrained backbones. The sample counts per input differ because they follow the respective Gaussian and Choice recipes. Table~\ref{tab:rlcd-design} summarizes the differences between the complete strategies.

\begin{table}[htbp]
\centering\small
\caption{RLCD recipes evaluated on Omni-JEV. Initialization, data order, input representation, and optimizer-update budget are shared. Categorical is our reproduction of Qwen-Audio-Jev's Choice algorithm, applied to Omni-JEV.}
\label{tab:rlcd-design}
\begin{tabular}{@{}>{\raggedright\arraybackslash}p{0.22\linewidth}>{\raggedright\arraybackslash}p{0.36\linewidth}>{\raggedright\arraybackslash}p{0.36\linewidth}@{}}
\toprule
Setting & Gaussian RLCD & Categorical RLCD\\
\midrule
Exploration unit & Centered logit vector & Discrete candidate position\\
Target score & Log plus spherical & Negative Brier\\
Samples per input & $M=16$ & $M=32$\\
Exploration control & $\sigma=0.1$ & $\eta=0.05$\\
Additional coefficient & Spherical weight $\beta=0.75$ & Ordinal weight $\lambda=0$\\
\bottomrule
\end{tabular}
\end{table}

\subsection{Metrics}
We report accuracy and macro-F1 for classification, and Brier error and negative log-likelihood (NLL) for probability quality:
\begin{equation}
 B=\frac1N\sum_i\sum_k(p_{ik}-t_{ik})^2,\qquad
 \mathrm{NLL}=-\frac1N\sum_i\log p_{i,y_i}.
\end{equation}
Brier values are displayed as $10^4B$. Macro-F1 averages actions with positive target support: five for audio and four for text, which has no \texttt{recover} examples. Modalities are reported separately. Pooled scores are recomputed after combining all test predictions, with macro-F1 over the five supported actions; this is not an average of modality-specific F1. It describes the corpus mixture rather than balanced cross-modal performance. Support and source/language slices appear in Appendices~\ref{app:support} and~\ref{app:slices}.

\subsection{External models and input conditions}\label{sec:external-protocol}
Qwen3-Omni-30B-A3B-Instruct, MiniCPM-o 4.5, MOSS-Audio-4B-Instruct, and Step-Audio-2-mini use all 34,860 tests~\cite{qwen3omnireport2025,qwen3omnicheckpoint,minicpmo2026,mossaudio2026,stepaudio22025}; FireRedAudio and AudioJev use all 24,287 audio tests~\cite{fireredaudio2026,audiojev2026}. Qwen-Audio-3.1-Realtime uses a fixed 2,000-example subset: 1,349 audio and 651 text~\cite{qwenrealtime2026}. Stratification by language, modality, and action among complete inputs no longer than 120 seconds gives this subset a different sampling distribution from the full test.

External models receive available user audio, text context, assistant state, and the same action definitions and candidate order. They are zero-shot with user-only audio; Omni-JEV is task-adapted and can also use assistant-reference audio. Table~\ref{tab:external-comparison} therefore compares task-level performance under these conditions.

Local generative baselines use greedy decoding, a 64-token output limit, and a parser accepting exact candidate symbols or action names; AudioJev uses its native numeric readout. Invalid outputs and unresolved requests remain errors in the denominator. Realtime retries preserve fixed IDs and initially successful answers (Appendix~\ref{app:api-recovery}).

\section{Results}\label{sec:results}
\subsection{RQ1: Text and speech action recognition}
\begin{figure}[t]
\centering
\includegraphics[width=\linewidth]{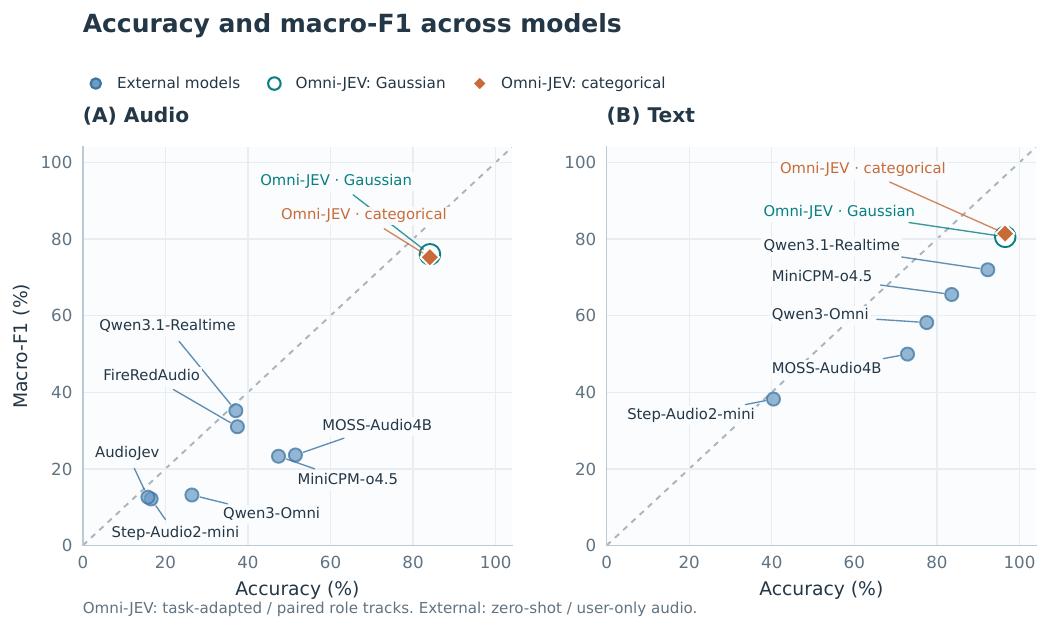}
\caption{Audio (A) and text (B) accuracy and macro-F1 from Table~\ref{tab:external-comparison}. Blue: external models; teal: Gaussian; orange: categorical. The diagonal marks F1 = ACC. Qwen3.1-Realtime uses the fixed 2,000-example subset and is not paired with full-scope systems.}
\label{fig:external-models}
\end{figure}

Table~\ref{tab:external-comparison} and Figure~\ref{fig:external-models} compare the models on the shared task. Gaussian Omni-JEV achieves 84.08\% audio accuracy and 76.00\% macro-F1, with 96.55\% text accuracy and 80.56\% macro-F1. Both adapted endpoints score higher than the external full-scope systems on these aggregates under this protocol. Because adaptation and available inputs differ, this is a task-level comparison rather than evidence of an equal-input architectural advantage.

\begin{table}[htbp]
\centering\scriptsize
\caption{Task-level comparison: the upper block uses the Realtime subset, the lower each model\textquotesingle s full applicable scope. External systems use zero-shot user-only audio; Omni-JEV is adapted and can also use assistant-reference audio. Failures remain in denominators. Bold marks modality-wise best full-scope scores.}
\label{tab:external-comparison}
\resizebox{\linewidth}{!}{%
\begin{tabular}{llrr|rr|rr}
\toprule
Model & Scope & $N$ & Fail/invalid & Audio ACC $\uparrow$ & Audio Macro-F1 $\uparrow$ & Text ACC $\uparrow$ & Text Macro-F1 $\uparrow$\\
\midrule
\multicolumn{8}{l}{\textit{Qwen3.1-Realtime API only; 2,000-row budget-limited subset}}\\
Qwen3.1-Realtime API & 2k API & 2,000 & 0/0 & 37.06 & 35.18 & 92.32 & 71.97\\
\multicolumn{8}{l}{\textit{Full frozen evaluation; AudioJev is audio-only}}\\
Qwen3-Omni & full test & 34,860 & 0/0 & 26.43 & 13.17 & 77.53 & 58.18\\
MiniCPM-o4.5 & full test & 34,860 & 0/0 & 47.40 & 23.28 & 83.58 & 65.51\\
MOSS-Audio4B & full test & 34,860 & 0/0 & 51.50 & 23.61 & 72.87 & 49.95\\
Step-Audio2-mini & full test & 34,860 & 0/12941 & 16.54 & 12.13 & 40.40 & 38.21\\
FireRedAudio & full audio & 24,287 & 0/0 & 37.44 & 30.99 & -- & --\\
AudioJev & full audio & 24,287 & 0/0 & 15.76 & 12.59 & -- & --\\
Omni-JEV Gaussian RLCD & full test & 34,860 & 0/0 & \textbf{84.08} & \textbf{76.00} & \textbf{96.55} & 80.56\\
Omni-JEV categorical RLCD & full test & 34,860 & 0/0 & 84.04 & 75.23 & 96.51 & \textbf{81.41}\\
\bottomrule
\end{tabular}
}

\end{table}

The task-level ranking includes differences in adaptation, role evidence, output format, and backbone. Step-Audio2-mini's 12,941 parser-invalid answers count as failures, so its score also reflects compliance with the decision format. Realtime uses the fixed subset specified in the table.

\subsection{RQ2: Comparing RLCD continuation strategies}
\begin{figure}[t]
\centering
\includegraphics[width=\linewidth]{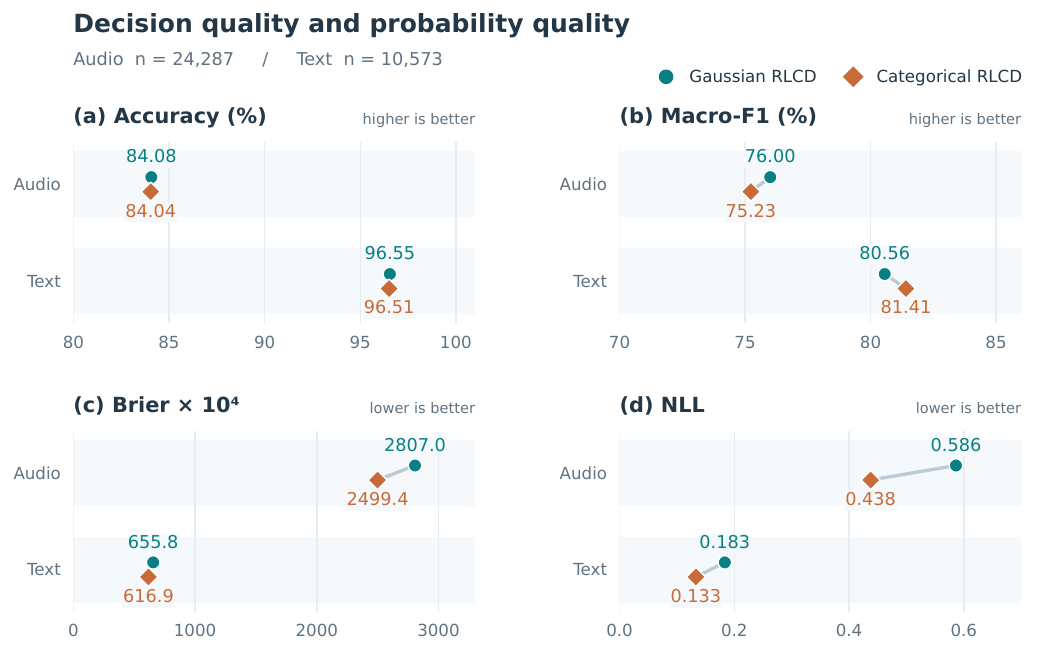}
\caption{Classification and probability quality after matched RLCD continuation. Circles denote Gaussian and diamonds categorical. ACC and macro-F1 axes are enlarged rather than zero-based; panels have separate ranges. Text macro-F1 covers four supported actions and audio five. Points are single-run estimates.}
\label{fig:results}
\end{figure}

Table~\ref{tab:main-results} and Figure~\ref{fig:results} compare continuations under Section~\ref{sec:rlcd-protocol}'s matched conditions.

\paragraph{Action classification.}
On audio, Gaussian has a macro-F1 advantage of 0.77 percentage points; on text, categorical leads by 0.85 points. Accuracy is nearly equal: Gaussian has eight more correct audio predictions and four more correct text predictions. These differences describe the prescribed single-seed runs.

\begin{table}[htbp]
\centering\small
\caption{Full-test RLCD comparison. ACC and macro-F1 are percentages. Both models use SFT$\rightarrow$RLCD; categorical reproduces Qwen-Audio-Jev\textquotesingle s Choice on Omni-JEV. Modality-wise best scores, including displayed ties, are bold. Pooled rows recompute metrics over the combined test predictions.}
\label{tab:main-results}
\begin{tabular}{llrrrrr}
\toprule
Modality & RLCD & $N$ & ACC $\uparrow$ & Macro-F1 $\uparrow$ & Brier $\times10^4\downarrow$ & NLL $\downarrow$\\
\midrule
Audio & Gaussian & 24,287 & \textbf{84.08} & \textbf{76.00} & 2806.99 & 0.5863\\
Audio & Categorical & 24,287 & 84.04 & 75.23 & \textbf{2499.43} & \textbf{0.4376}\\
Text & Gaussian & 10,573 & \textbf{96.55} & 80.56 & 655.79 & 0.1834\\
Text & Categorical & 10,573 & 96.51 & \textbf{81.41} & \textbf{616.87} & \textbf{0.1331}\\
Pooled & Gaussian & 34,860 & \textbf{87.86} & \textbf{80.93} & 2154.54 & 0.4641\\
Pooled & Categorical & 34,860 & 87.83 & 80.32 & \textbf{1928.46} & \textbf{0.3452}\\
\bottomrule
\end{tabular}

\end{table}

\paragraph{Probability quality.}
Categorical also has lower Brier error and NLL on both modalities. Relative to Gaussian, Brier error is lower by 10.96\% on audio and 5.93\% on text, computed from unrounded aggregates. Accuracy depends on the largest-probability candidate, Brier on the full vector, and NLL on the target probability. Their different rankings show that similar action agreement can coexist with different probability quality; neither continuation dominates across metrics. This distinction matters for a decision interface that exposes probabilities as well as a selected action: aggregate classification scores alone do not describe the reliability of those probability assignments. The present comparison measures both properties without establishing a preferred strategy for a particular controller.

\subsection{RQ3: Source and language variation}
Natural action frequencies are uneven: audio has 15,617 \texttt{listen} examples but only 17 \texttt{recover} examples; text has 81 \texttt{listen} examples and no \texttt{recover} examples (Appendix~\ref{app:support}). Macro-F1 is sensitive to rare-action errors, whereas accuracy reflects dominant actions.

Appendix~\ref{app:slices} reports Gaussian audio accuracy of 81.90\% on Chinese and 94.11\% on English, with 40 nonlinguistic examples. Source, action composition, translation, and channel quality vary with language: SmoothConv contributes 15,003 audio tests and FD 184. These slices locate heterogeneity without isolating a language effect.

The source scores also differ in what they measure. Gaussian reaches 98.61\% accuracy and 98.54\% macro-F1 on Mini Speech Commands, versus 78.41\% and 55.26\% on SmoothConv. The former maps isolated words to actions by convention; the latter derives targets from conversational role activity. MInDS-14 and ESC-10 each use a single scenario-assigned action. Thus the aggregate combines lexical conventions, added scenarios, and conversational prediction rather than a uniform test of playback control. The construction table in Appendix~\ref{app:annotation} and source slices should be read together; their scores alone neither establish reliable interruption handling nor identify the cause of the source differences.

\section{Discussion and Limitations}
The common action target makes a distinction that matters for response control: recognizing a contribution and deciding its effect on an existing response are related but different predictions. Turn states, semantic interruptions, and next-speaker behavior provide evidence for that decision~\cite{fastturn2026,x2turn2026,sidbench2026,audiojev2026}. Omni-JEV expresses it through one readout for text and role-indexed speech. Direct speech retains potential prosodic and temporal cues~\cite{fastturn2026}, while assistant-reference spans provide the response context. The offline results characterize learning under this formulation; they do not isolate gains from waveform input, separate roles, or STS initialization. External baselines also differ in adaptation and available inputs, so the task-level ranking does not establish an equal-input architectural advantage.

The RLCD results distinguish two properties of the output. Similar argmax accuracy can coexist with different error distributions across actions and different probability assignments. Gaussian has higher audio macro-F1; categorical has higher text macro-F1 and lower Brier error and NLL in both modalities. Categorical directly targets negative Brier, making its lower Brier error consistent with that objective. However, rewards, exploration, and estimators change jointly, so the comparison concerns complete strategies. Proper scoring principles~\cite{proper2007} concern ideal distributions; they do not guarantee calibration after finite-sample optimization. The result does not establish a universally preferable continuation or calibration within every response state, and does not estimate gains over supervised fine-tuning alone.

Generalization depends on the weak supervision policy and the available evidence. Lexical conventions and added scenarios coexist with conversational targets; isolated commands therefore do not test compatibility with actual playback state. Source/action imbalance, synthesized speech, sparse recovery, and unknown semantic duplicates or pretraining exposure limit the aggregate interpretation. Text and audio have different source mixtures, so their accuracy gap does not compare transcripts with waveforms. Mono role masks also retain mixed voices during overlap. The results use one training seed, and the test set has been inspected during diagnosis. Interactive benchmarks~\cite{pace2026,interactivity2026,fdbenchv2_2026,humdial2026} address complementary behavior; synchronization, acoustic echo robustness, latency, and multi-turn utility remain unmeasured here.

\section{Conclusion}
Omni-JEV formulates collected-segment decisions as a state-conditioned action distribution shared by text and role-indexed speech. On the fixed weak-label test partition, Gaussian Omni-JEV reaches 84.08\% audio accuracy and 96.55\% text accuracy; categorical RLCD obtains lower Brier error and NLL, while Gaussian has slightly higher audio macro-F1. These results answer the study question in a bounded way: one candidate-distribution interface can express the same endpoint-action task across text and role-indexed speech, while the preferred RLCD continuation depends on whether classification balance or probability quality is prioritized. For a full-duplex controller, the interface can provide an action and its distribution after a collected segment; closed-loop latency, acoustic robustness, and multi-turn utility remain to be established.

\par\medskip
\begingroup\raggedright\sloppy
\bibliographystyle{plain}\bibliography{references}

\begin{thebibliography}{10}

\bibitem{qwenrealtime2026}
{Alibaba Token Foundry, Alibaba Group}.
\newblock {Qwen-Audio-3.1-Realtime}: Towards reliable agentic voice
  interaction.
\newblock \href{https://arxiv.org/abs/2609.25176}{arXiv:2609.25176}, 2026.

\bibitem{funaudiollm2024}
Keyu An, Qian Chen, Chong Deng, Zhihao Du, et~al.
\newblock {FunAudioLLM}: Voice understanding and generation foundation models
  for natural interaction between humans and {LLMs}.
\newblock \href{https://arxiv.org/abs/2407.04051}{arXiv:2407.04051}, 2024.

\bibitem{minicpmo2026}
Junbo Cui, Bokai Xu, Chongyi Wang, Tianyu Yu, et~al.
\newblock {MiniCPM-o 4.5}: Towards real-time full-duplex omni-modal
  interaction.
\newblock \href{https://arxiv.org/abs/2604.27393}{arXiv:2604.27393}, 2026.

\bibitem{moshi2024}
Alexandre D{\'e}fossez, Laurent Mazar{\'e}, Manu Orsini, Am{\'e}lie Royer,
  Patrick P{\'e}rez, Herv{\'e} J{\'e}gou, Edouard Grave, and Neil Zeghidour.
\newblock Moshi: a speech-text foundation model for real-time dialogue, 2024.

\bibitem{vap2022}
Erik Ekstedt and Gabriel Skantze.
\newblock Voice activity projection: Self-supervised learning of turn-taking
  events, 2022.

\bibitem{massive}
Jack FitzGerald, Christopher Hench, Charith Peris, Scott Mackie, et~al.
\newblock {MASSIVE}: A 1m-example multilingual natural language understanding
  dataset with 51 typologically-diverse languages.
\newblock In {\em Proceedings of the 61st Annual Meeting of the Association for
  Computational Linguistics (Volume 1: Long Papers)}, pages 4277--4302, 2023.

\bibitem{x2turn2026}
Kaiqi Fu, Rime Wen, Altman Lin, Shawn Qin, Roy Gan, Hao Wang, and Qian Wang.
\newblock X2-turn: Frame-synchronous dual-head modeling for joint streaming asr
  and turn state prediction, 2026.

\bibitem{sensevoicesource}
{FunAudioLLM Team}.
\newblock {SenseVoice}: Multilingual speech understanding.
\newblock Official repository, 2024.

\bibitem{minds14}
Daniela Gerz, Pei-Hao Su, Razvan Kusztos, Avishek Mondal, et~al.
\newblock Multilingual and cross-lingual intent detection from spoken data.
\newblock In {\em Proceedings of the 2021 Conference on Empirical Methods in
  Natural Language Processing}, pages 7468--7475, 2021.

\bibitem{proper2007}
Tilmann Gneiting and Adrian~E. Raftery.
\newblock Strictly proper scoring rules, prediction, and estimation.
\newblock {\em Journal of the American Statistical Association},
  102(477):359--378, 2007.

\bibitem{minimindo2026}
Jingyao Gong.
\newblock {MiniMind-O} technical report: An open small-scale speech-native omni
  model.
\newblock \href{https://arxiv.org/abs/2605.03937}{arXiv:2605.03937}, 2026.

\bibitem{salmduplex2025}
Ke~Hu, Ehsan Hosseini-Asl, Chen Chen, Edresson Casanova, et~al.
\newblock {SALM-Duplex}: Efficient and direct duplex modeling for
  speech-to-speech language model.
\newblock \href{https://arxiv.org/abs/2505.15670}{arXiv:2505.15670}, 2025.

\bibitem{fdbenchv2_2026}
Guan-Ting Lin, Shih-Yun~Shan Kuan, Jiatong Shi, Kai-Wei Chang, et~al.
\newblock {Full-Duplex-Bench-v2}: A multi-turn evaluation framework for duplex
  dialogue systems with an automated examiner.
\newblock \href{https://arxiv.org/abs/2510.07838}{arXiv:2510.07838}, 2025.
\newblock Revised April 2026; accepted to ACL 2026, as recorded by the authors.

\bibitem{fdbench2025}
Guan-Ting Lin, Jiachen Lian, Tingle Li, Qirui Wang, et~al.
\newblock {Full-Duplex-Bench}: A benchmark to evaluate full-duplex spoken
  dialogue models on turn-taking capabilities.
\newblock \href{https://arxiv.org/abs/2503.04721}{arXiv:2503.04721}, 2025.
\newblock Accepted to ASRU 2025, as recorded by the authors.

\bibitem{routingduplex2026}
Hui Lu, Xueyuan Chen, Huimeng Wang, Shuhai Peng, Shiyin Kang, Xixin Wu, and
  Zhiyong Wu.
\newblock How should {LLMs} listen while speaking? a study of user-stream
  routing in full-duplex spoken dialogue.
\newblock \href{https://arxiv.org/abs/2605.10199}{arXiv:2605.10199}, 2026.

\bibitem{audiojev2026}
Sihan Lv, Zhen Li, Zhiqi Cao, Jinshan Zhang, Ying Li, Meng Xi, and Jianwei Yin.
\newblock {AudioJev}: Direct audio decisions with order-calibrated
  probabilities, 2026.

\bibitem{layasource}
{NandhaKishorM and Laya contributors}.
\newblock {Laya}: Gaussian-logit {RLCD} training implementation.
\newblock Pinned source repository, 2026.
\newblock Audited commit 8a6e1328cce2460a0e5aa348ad465bb1b5821cd2; source
  provenance retained in the project archive.

\bibitem{banking77}
I\~nigo Casanueva, Tadas Tem\v{c}inas, Daniela Gerz, Matthew Henderson, and
  Ivan Vuli\'{c}.
\newblock Efficient intent detection with dual sentence encoders.
\newblock In {\em Proceedings of the 2nd Workshop on Natural Language
  Processing for Conversational AI}, pages 38--45, 2020.

\bibitem{interactivity2026}
Atsumoto Ohashi, Neil Zeghidour, Alexandre D{\'e}fossez, and Eugene Kharitonov.
\newblock Multi-faceted interactivity alignment in full-duplex speech models.
\newblock \href{https://arxiv.org/abs/2606.11167}{arXiv:2606.11167}, 2026.

\bibitem{legacytextdata}
{Omni-JEV project}.
\newblock Legacy turn-text corpus and derived five-action annotation manifest.
\newblock Unpublished project data record, 2026.
\newblock Internal unpublished source, not a separately published benchmark;
  underlying text ownership and external publication permission are not
  established by the current ledger.

\bibitem{mossaudio2026}
{OpenMOSS Team}.
\newblock Moss-audio: Open-source audio understanding model.
\newblock GitHub repository and released model weights, 2026.
\newblock Release notes dated April 13, 2026; Apache-2.0 model license.

\bibitem{escsource}
Karol~J. Piczak.
\newblock {ESC}: Dataset for environmental sound classification.
\newblock Official dataset repository, 2015.

\bibitem{smoothconvsource}
{qualialabsAI}.
\newblock {SmoothConv}: Conversation dataset.
\newblock Dataset repository.
\newblock Source identity retained in the archived corpus and migration
  manifest; dataset card declares CC BY-NC 4.0 and academic/research use.

\bibitem{qwen3omnicheckpoint}
{Qwen Team}.
\newblock Qwen3-omni-30b-a3b-instruct model card and checkpoint.
\newblock ModelScope model repository.
\newblock Local snapshot used for the October 23, 2026 evaluation.

\bibitem{rsisource}
{Shanghua-Gao and RSI-Jev contributors}.
\newblock {RSI-Jev}: Calibrated decision training and estimator diagnostics.
\newblock Pinned source repository, 2026.
\newblock Audited commit 7066d5f037fefe11ab50a4caf6f6ac16a89a2304.

\bibitem{fireredaudio2026}
Feiyu Shen, Fenglong Xie, Junjie Li, Kun Xie, et~al.
\newblock {FireRedAudio}: A general-purpose audio language model with decoupled
  continuous representations for understanding and generation.
\newblock \href{https://arxiv.org/abs/2608.24168}{arXiv:2608.24168}, 2026.

\bibitem{wakeupsource}
{TCL Research Europe}.
\newblock {Intelligent Wakeup}: Dataset card and source repository.
\newblock Hugging Face dataset repository.
\newblock Original dataset card archived in the project; CC BY-NC 4.0 data;
  source used in October 2026.

\bibitem{qwenaudiojev2026}
{Token Foundry, Alibaba Group}.
\newblock Qwen-audio-jev: A unified interface for audio decisions, 2026.

\bibitem{humdial2026}
Chengyou Wang, Hongfei Xue, Guojian Li, Zhixian Zhao, et~al.
\newblock Full-duplex interaction in spoken dialogue systems: A comprehensive
  study from the {ICASSP 2026 HumDial} challenge.
\newblock \href{https://arxiv.org/abs/2604.21406}{arXiv:2604.21406}, 2026.

\bibitem{fastturn2026}
Chengyou Wang, Hongfei Xue, Mingchen Shao, Chunjiang He, et~al.
\newblock {FastTurn}: Unifying acoustic and streaming semantic cues for
  low-latency and robust turn detection.
\newblock \href{https://arxiv.org/abs/2604.01897}{arXiv:2604.01897}, 2026.

\bibitem{pace2026}
Shibo Wang, Zicheng Zhang, Libo Wang, and Junfeng Ma.
\newblock Pace: A playback-aligned context engine for llm-based full-duplex
  voice dialogue, 2026.

\bibitem{freezeomni2024}
Xiong Wang, Yangze Li, Chaoyou Fu, Yunhang Shen, et~al.
\newblock {Freeze-Omni}: A smart and low latency speech-to-speech dialogue
  model with frozen {LLM}.
\newblock \href{https://arxiv.org/abs/2411.00774}{arXiv:2411.00774}, 2024.

\bibitem{speechcommands}
Pete Warden.
\newblock Speech commands: A dataset for limited-vocabulary speech recognition,
  2018.

\bibitem{reinforce1992}
Ronald~J. Williams.
\newblock Simple statistical gradient-following algorithms for connectionist
  reinforcement learning.
\newblock {\em Machine Learning}, 8:229--256, 1992.

\bibitem{stepaudio22025}
Boyong Wu, Chao Yan, Chen Hu, Cheng Yi, et~al.
\newblock {Step-Audio 2} technical report.
\newblock \href{https://arxiv.org/abs/2507.16632}{arXiv:2507.16632}, 2025.
\newblock Evaluated checkpoint: Step-Audio-2-mini; inference code at
  \url{https://github.com/stepfun-ai/Step-Audio2}.

\bibitem{sidbench2026}
Kangxiang Xia, Bingshen Mu, Xian Shi, Jin Xu, and Lei Xie.
\newblock Semantic-aware interruption detection in spoken dialogue systems:
  Benchmark, metric, and model, 2026.

\bibitem{qwen3omnireport2025}
Jin Xu, Zhifang Guo, Hangrui Hu, Yunfei Chu, et~al.
\newblock {Qwen3-Omni} technical report.
\newblock \href{https://arxiv.org/abs/2509.17765}{arXiv:2509.17765}, 2025.

\bibitem{qwen3report2025}
An~Yang, Anfeng Li, Baosong Yang, Beichen Zhang, et~al.
\newblock {Qwen3} technical report.
\newblock \href{https://arxiv.org/abs/2505.09388}{arXiv:2505.09388}, 2025.

\bibitem{fdv2source}
{zetianli}.
\newblock {FD\_data\_v2}: Synthetic full-duplex conversation dataset.
\newblock Hugging Face dataset repository.
\newblock Original conversation source; upstream card declares Apache-2.0;
  derived action windows are described separately.

\bibitem{fdv3source}
{zetianli}.
\newblock {FD\_data\_v3}: Synthetic full-duplex conversation dataset.
\newblock Hugging Face dataset repository.
\newblock Original conversation source; license not independently verified in
  the retained archive; derived action windows are described separately.

\bibitem{duplexsla2026}
Haoyang Zhang, Jun Chen, Donghang Wu, Yuxin Li, et~al.
\newblock {DuplexSLA}: A full-duplex spoken language model with synchronized
  speech, language, and action.
\newblock \href{https://arxiv.org/abs/2605.20755}{arXiv:2605.20755}, 2026.

\end{thebibliography}
\endgroup
\clearpage
\appendix
\section{Corpus Support and Imbalance}\label{app:support}
Table~\ref{tab:support} gives action support within each modality. Matching training/test proportions preserves the natural imbalance; it does not balance classes. In particular, text is dominated by \texttt{start}/\texttt{stop}, whereas audio is dominated by \texttt{listen}. The natural-frequency protocol uses neither class weights nor balanced replacement sampling.

\begin{table}[H]
\centering\small
\caption{Training and test action counts in the fixed corpus partition.}
\label{tab:support}
\begin{tabular}{llrr}
\toprule
Modality & Action & Train & Test\\
\midrule
audio & \texttt{listen} & 138,507 & 15,617\\
audio & \texttt{start} & 52,771 & 5,764\\
audio & \texttt{continue} & 21,545 & 2,373\\
audio & \texttt{stop} & 4,720 & 516\\
audio & \texttt{recover} & 156 & 17\\
text & \texttt{listen} & 737 & 81\\
text & \texttt{start} & 49,482 & 5,462\\
text & \texttt{continue} & 14,590 & 1,610\\
text & \texttt{stop} & 30,978 & 3,420\\
text & \texttt{recover} & 0 & 0\\
\bottomrule
\end{tabular}

\end{table}

\paragraph{Eligibility.}
The frozen corpus excludes 74,516 English Wakeup rows whose verified input overlay does not identify a role-confirmed external-user VAD pause. This is an observation-boundary eligibility filter, not a model-error or annotation-confidence filter. It also excludes 1,544 exact duplicates, and 24 conflicting identical inputs. Planned additional ASR/TTS augmentation and demonstration-only generations were not used in these experiments. Corpus counts refer only to eligible examples in the fixed training/test partition.

\section{Annotation and Prefix Provenance}\label{app:annotation}
The common action vocabulary does not imply identical observation or annotation mechanisms across sources. We retain the existing derived action labels and source links during corpus assembly, rather than relabeling every source with one teacher.

\paragraph{Wakeup.}
Original speaker identities assign Sigma to the assistant role and the remaining speakers to external evidence. The English corpus admits only externally confirmed user-pause records after the input-state repair. Its native-role-block anchor includes the latest observed substantive assistant block and its preceding external block; pending interruptions or required earlier evidence can extend the retained history. The Chinese version reuses source semantic targets, aligns them to independently detected Chinese boundaries, and supplements changed cases. Both versions retain the native-role-block prefix policy. Native or synthetic speaker-time masks supply the two inputs and cannot recover clean sources from mono overlap. The cropped-prefix construction reuses full-prefix teacher labels; those labels were not all independently rejudged on the cropped input.

\paragraph{FD.}
The final prefix policy uses an observed assistant main turn and its preceding user block; the English and Chinese versions preserve synchronized role tracks. The saved user-pause annotation pipeline applies Silero VAD to the roles separately and requests decisions only after a confirmed user speech end. Its configuration uses 32-ms frames, onset/release thresholds $0.5/0.35$, at least 100 ms of speech, and 500 ms of silence for end confirmation. SenseVoice transcripts are computed from observed segments, and DeepSeek-nothink receives those transcripts with replayed response state and the known policy. Original action labels and future utterances are excluded from those teacher requests. The larger Chinese expansion reuses source decisions after boundary realignment; it is not a second independent annotation of every utterance. These settings describe the archived FD pipeline, not a verified common VAD version for every source.

\paragraph{SmoothConv.}
Here self denotes the assistant role and external the user role. Native external segment ends supply decision points without another VAD pass. The rule priority is \texttt{stop}, then \texttt{continue}, then \texttt{recover}/\texttt{start}, otherwise \texttt{listen}. A stop requires a self turn beginning at least 0.3 s earlier, overlap of at least 0.1 s, external speech continuing at least 0.1 s after self, and either an incomplete self turn or an explicit interruption marker; ordinary backchannels are excluded. One self turn cannot trigger repeated stops. Continued self speech at the endpoint yields \texttt{continue}. Otherwise, a self main turn within a 5-s annotation horizon, preceding the next external main turn with no external main turn still active, yields \texttt{recover} if a stop remains pending and \texttt{start} otherwise. A stop sets the pending flag, listen retains it, and start/recover clears it; an observed intervening self-resumption also clears stale pending state. The 5-s horizon is annotation lookahead, not an inference delay. Prefixes use the segment beginning of the preceding rule-labeled start, so their anchor trajectory depends on rule supervision rather than deployed model predictions.

\paragraph{Serialized state and history.}
The retained \texttt{assistant\_state} keys are \texttt{value} and, when supplied, \nolinkurl{pending_interrupted_response}, \nolinkurl{resumable_response}, and \nolinkurl{self_spoke_during_collected_external_segment}. In SmoothConv, \texttt{value} is reconstructed from self activity at the endpoint and prior pending state, before updating with the current rule action; the interval-speech flag separately records overlap during the collected external segment. Wakeup repairs recover the source observation at the same cutoff, not device telemetry. Completed source text before the audio window can be retained under \nolinkurl{completed_history_before_audio_window}; current-window and partial transcripts are excluded. The prompt serializes \texttt{text}, \texttt{assistant\_state}, \texttt{policy}, and \texttt{scenario} as context, while audio spans and candidate options are supplied separately. These fields do not constitute an independently validated state estimator.

\paragraph{Short-form sources and translations.}
Table~\ref{tab:construction} distinguishes source evidence, assigned state, and the procedure that produces the weak action target. The text teacher uses the saved \texttt{duplex-five-action-}\allowbreak\texttt{pilot-v2-strict} policy: idle, unfinished or non-addressed input yields \texttt{listen}; a completed request or an appropriate clarification yields \texttt{start}. Under task-given playback, a non-disruptive contribution yields \texttt{continue}, whereas a substantive correction or new request yields \texttt{stop}. The original domain-intent ID is excluded from the teacher view. Saved DeepSeek responses retain requested and returned model identifiers; they do not identify one immutable teacher checkpoint. Mini Speech Commands is a separate lexical convention: its unknown state does not verify compatibility of \texttt{continue} or \texttt{stop} with actual playback. No assistant history is synthesized for these isolated clips. Source links join translations before partitioning.

\begingroup\footnotesize
\setlength{\tabcolsep}{4pt}
\renewcommand{\arraystretch}{1.16}
\begin{longtable}{@{}>{\raggedright\arraybackslash}p{0.14\linewidth}>{\raggedright\arraybackslash}p{0.25\linewidth}>{\raggedright\arraybackslash}p{0.30\linewidth}>{\raggedright\arraybackslash}p{0.25\linewidth}@{}}
\caption{From source material to weak action targets. Observed conversational state, task-given state, and added scenarios are distinguished. Label reuse preserves the source decision under the intended translation; it is not an independent semantic validation.}\label{tab:construction}\\
\toprule
Source & Source unit and state & Action construction & Bilingual handling\\\midrule
\endfirsthead
\toprule
Source & Source unit and state & Action construction & Bilingual handling\\\midrule
\endhead
\bottomrule\endfoot
BANKING77 & Bank-intent text; idle/no-resumable state added; no assistant history. & Saved DeepSeek text policy described above; the original intent category is excluded. & Translate text fields; retain assigned state and action.\\
MASSIVE & Intent text; the same added idle scenario, without a prior assistant turn. & The same text teacher distinguishes a completed contribution from unfinished or non-addressed input. & Translate text fields; retain assigned state and action.\\
Legacy turn text & Assistant/user text pair; playing state supplied by the source task instruction. & The text teacher rejudges the pair under the five-action policy; native three-class targets are not directly projected. & Translate both text roles; retain state and derived action.\\
Mini Speech Commands & Isolated word recording and its word class; context absent and assistant state unknown. & Fixed convention: stop/no $\to$ \texttt{stop}; yes/go $\to$ \texttt{continue}; up/down/left/right $\to$ \texttt{start}. Playback compatibility is unverified. & Chinese TTS of the translated word; copy the action without VAD or ASR.\\
MInDS-14 & User utterance; idle/no-resumable state and completed direct request assumed at clip end. & Scenario rule assigns \texttt{start}; native bank intent and transcript are not consulted. & English TTS uses the source-provided English transcription; copy the action without VAD.\\
ESC-10 & Environmental or nonlinguistic clip; idle/no-resumable scenario added, with no sound-identification or alarm task. & Under that scenario, assign \texttt{listen} to every included clip. & Nonlinguistic; no translated version is required.\\
Wakeup & Speaker-timed conversation; original identities and role blocks determine observed evidence. & Existing semantic targets plus repaired user-pause eligibility; detailed boundary and role rules above. & Reuse source targets after independent Chinese boundary alignment; supplement changed cases.\\
FD & Synthesized conversation with synchronized roles; replayed state at a confirmed user pause. & Saved causal ASR plus DeepSeek-nothink decisions; original actions and future utterances excluded from teacher input. & Align Chinese boundaries separately; the expansion reuses source decisions after realignment.\\
SmoothConv & Native role/channel segments; external segment ends and observed role activity define the sample. & Ordered temporal rules above, with later behavior available only for offline annotation. & Chinese recordings; no translated counterpart used in these experiments.\\
\end{longtable}
\endgroup

\paragraph{A saved input--target example.}
A MASSIVE English variant contains the text ``Weibo'', with assistant state \texttt{idle}, no audio or assistant-text history, empty additional policy and scenario fields, and the five candidates \texttt{listen}, \texttt{start}, \texttt{continue}, \texttt{stop}, and \texttt{recover}. Its saved weak target is \texttt{start}: the teacher interprets the short contribution as permitting a clarifying response. The idle/no-resumable condition is an added annotation scenario, not a measured property of the source utterance. The native intent \texttt{social\_query} is retained only as provenance, outside the model input. This example documents the conversion rather than independently validating the target.

\paragraph{Documentation boundary.}
The frozen corpus incorporates the existing input repairs without changing audio or targets. The source records retain policy identifiers, but the complete per-example teacher revisions and exact VAD/ASR/translation model snapshots are not packaged in this manuscript. Mechanical boundary checks do not replace independent semantic adjudication, particularly when labels use later behavior or when a translated boundary merges or separates source segments.

\section{Source and Language Slices}\label{app:slices}
The tables below are descriptive slices of the same fixed test predictions, not additional experiments. G denotes Gaussian and C categorical RLCD. ACC and F1 are percentages; each slice's F1 averages its supported target classes. This can make high F1 on single-action sources misleading without the corresponding support table.

\begin{table}[H]
\centering\small
\caption{Source-specific test results. Bold marks the better recipe for each metric within a source; ties at displayed precision are retained.}
\begin{tabular}{lrrrrr}
\toprule
Source & $N$ & G ACC & G F1 & C ACC & C F1\\
\midrule
BANKING77 & 2,509 & \textbf{99.64} & \textbf{49.91} & 99.60 & 49.90\\
MASSIVE & 3,034 & \textbf{97.86} & 69.08 & 97.79 & \textbf{70.99}\\
Legacy turn text & 5,030 & \textbf{94.21} & \textbf{93.36} & 94.19 & 93.34\\
Mini Speech Commands & 1,580 & \textbf{98.61} & \textbf{98.54} & 98.42 & 98.31\\
MInDS-14 & 100 & \textbf{100.00} & \textbf{100.00} & \textbf{100.00} & \textbf{100.00}\\
ESC-10 & 40 & \textbf{100.00} & \textbf{100.00} & \textbf{100.00} & \textbf{100.00}\\
Wakeup & 7,380 & \textbf{92.56} & \textbf{69.13} & 92.43 & 67.25\\
FD & 184 & 69.02 & 51.43 & \textbf{69.57} & \textbf{57.54}\\
SmoothConv & 15,003 & 78.41 & \textbf{55.26} & \textbf{78.44} & 54.74\\
\bottomrule
\end{tabular}

\end{table}

\begin{table}[H]
\centering\small
\caption{Language-by-modality results. ``none'' is nonlinguistic audio. Bold marks the better recipe within each slice; ties at displayed precision are retained.}
\begin{tabular}{lrrrrr}
\toprule
Slice & $N$ & G ACC & G F1 & C ACC & C F1\\
\midrule
audio / zh & 19,968 & \textbf{81.90} & \textbf{73.34} & 81.86 & 72.43\\
audio / en & 4,279 & \textbf{94.11} & \textbf{93.31} & \textbf{94.11} & 93.17\\
audio / none & 40 & \textbf{100.00} & \textbf{100.00} & \textbf{100.00} & \textbf{100.00}\\
text / zh & 5,274 & \textbf{96.47} & 80.06 & 96.40 & \textbf{81.51}\\
text / en & 5,299 & \textbf{96.62} & 81.06 & \textbf{96.62} & \textbf{81.30}\\
\bottomrule
\end{tabular}

\end{table}

\paragraph{Limits of source aggregates.}
SmoothConv's source-level ACC and macro-F1 summarize agreement with the derived action policy; they do not by themselves establish reliable interruption handling. Short command sources use different target conventions and cannot supply that missing evidence. Annotation conventions and acoustic/input quality are plausible explanations for source differences, but the aggregates do not distinguish their effects.

\section{RLCD Objectives and Estimators}\label{app:rlcd-details}
\subsection{Relationship to prior Gaussian implementations}
The Gaussian-logit exploration, log-plus-spherical score, and density-gradient update are inherited from prior RLCD implementations~\cite{layasource,rsisource}. The archived Laya fine-tuning example already centers the perturbations and uses spherical weight $0.75$, group-mean-centered standardized advantages, and an auxiliary CE term. The RSI-Jev implementation uses a raw leave-one-out baseline with uncentered Gaussian noise. Our Omni-JEV recipe uses centered noise, unstandardized leave-one-out advantages, $M=16$, $\sigma=0.1$, and no auxiliary CE term. These choices specify the continuation evaluated here; we do not claim a new Gaussian estimator or a new exploration distribution from centering. The cited examples do not establish the training history of the published Laya weights.

\subsection{Gaussian surrogate and objective}
On the zero-sum subspace, the centered noise in Equation~\eqref{eq:gaussian} has covariance $\sigma^2P$. Softmax invariance gives $\operatorname{softmax}(z+\sigma P\xi)=\operatorname{softmax}(z+\sigma\xi)$, so centering changes neither the sampled probabilities nor their reward. The Gaussian density and score are understood on this subspace.

With sampled vectors and advantages detached, the implemented surrogate is
\begin{equation}\label{eq:surrogate}
 \mathcal L_G=-\frac1M\sum_r\operatorname{sg}(A_r)
 \left[-\frac{\|\widetilde z_r-z\|_2^2}{2\sigma^2}\right].
\end{equation}
Differentiating the density with respect to its center yields Equation~\eqref{eq:gaussian-gradient}. Because the leave-one-out baseline excludes sample $r$, its product with that sample's independent density score has zero expectation. Under regularity conditions, the estimate targets $J_\sigma(z)=\mathbb E_\epsilon R(\operatorname{softmax}(z+\epsilon),t)$. The advantage is equivalently $A_r=\frac{M}{M-1}(R(q_r,t)-\overline R)$, where $\overline R$ is the group mean.

\paragraph{Scoring properties.}
For an unsmoothed distribution $p$ with positive probabilities, using the continuous-extension convention $0\log0=0$ for the target, the idealized score satisfies
\begin{equation}\label{eq:proper}
 R(t,t)-R(p,t)=D_{\mathrm{KL}}(t\|p)
 +\beta\left(\|t\|_2-\frac{\langle t,p\rangle}{\|p\|_2}\right)\ge0.
\end{equation}
Nonnegativity follows from KL divergence and Cauchy--Schwarz, connecting the reward to proper scoring rules~\cite{proper2007}. The inequality concerns the ideal target score: it does not order the two trained models. Gaussian smoothing changes the optimized objective, and finite-sample updates and weak labels further separate this idealization from the inference distribution. Probability quality must therefore be measured rather than inferred from properness.

\paragraph{Estimator scope.}
The reward is differentiable, so the same smoothed objective also admits the reparameterized gradient $\nabla_z J_\sigma(z)=\mathbb E_\epsilon[\nabla_z R(\operatorname{softmax}(z+\epsilon),t)]$ under regularity conditions. Our experiment retains the score-function continuation from the prior Gaussian construction to compare it with categorical Choice on the action task. It does not establish a variance, efficiency, or optimization advantage over direct differentiation. The two-strategy comparison evaluates the implemented continuations without attributing their differences to the gradient estimator alone.

\subsection{Categorical Choice objective and update}
The reproduced Choice objective in Equation~\eqref{eq:categorical-objective} is estimated as follows. With $p=\operatorname{softmax}(z)$, candidates are sampled from $\mu=(1-\eta)p+\eta/K$ and weighted by $w_r=p(a_r)/\mu(a_r)$. The mixed proposal provides exploration; importance weighting corrects the expectation toward the model distribution. The sampled objective is
\begin{equation}\label{eq:categorical-reward}
 \widehat J_C=\frac2M\sum_rw_r t_{a_r}-1-
 \frac{1}{M(M-1)}\sum_{r\ne s}w_rw_s\mathbf1[a_r=a_s].
\end{equation}
The pairwise collision term estimates the squared probability norm. With $k_{rs}=\mathbf1[a_r=a_s]$ for $r\ne s$ and $k_{rr}=0$, the influence return, leave-one-out baseline, and surrogate loss are
\begin{equation}\label{eq:categorical-update}
\begin{aligned}
 I_r&=2t_{a_r}-\frac{2}{M-1}\sum_{s\ne r}w_s k_{rs},\\
 b_{-r}&=\frac{2}{M-1}\sum_{s\ne r}w_s t_{a_s}
 -\frac{2}{(M-1)(M-2)}
 \sum_{\substack{s\ne v\\s,v\ne r}}w_s w_v k_{sv},\\
 \mathcal L_C&=-\frac1M\sum_r
 \operatorname{sg}\!\left[w_r(I_r-b_{-r})\right]\log p(a_r).
\end{aligned}
\end{equation}
The pairwise baseline retains the factor of two in the reproduced algorithm. The proposal, importance weights, returns, and baseline are detached; only sampled $\log p(a_r)$ terms carry gradients. We use $M=32$, $\eta=0.05$, and no ordinal term ($\lambda=0$), following the reproduced Choice configuration.

\section{Evaluation Consistency and Availability}\label{app:evidence}
All main results use one frozen training/test partition. Earlier development subsets and supplementary evaluations are excluded from these results. The two RLCD continuations share the same supervised initialization, ordered training inputs, and update budget; their final checkpoints are determined by the training schedule rather than test performance. The partition and training protocol were fixed on October 23, 2026, the continuation training finished on October 27, and full endpoint verification finished on October 28. Test predictions were subsequently inspected for diagnosis and manuscript development. This is a fixed retrospective evaluation rather than a blind prospective benchmark.

\paragraph{Initialization specification.}\label{app:initialization}
The transferred frontend and Thinker come from our separately trained STS checkpoint built on MiniMind-O~\cite{minimindo2026}. Its source identifier is \texttt{s2s\_all/model\_epoch\_003}; the action-training asset manifest and supervised-stage initialization record agree on both weight files. The frozen copy has a Qwen3 decoder with 28 layers, width 1024, intermediate width 3072, head dimension 128, 16 attention heads, eight key/value heads, vocabulary size 151,936, tied word embeddings, and a 40,960-position limit. These are the actual starting-model settings. The available action-experiment records do not provide a complete STS alignment curriculum, loss specification, pretraining-data inventory, or exposure budget, so the STS stage is not independently reproducible from this manuscript alone. Its initialization is held fixed in both RLCD arms; those experiments do not estimate the benefit of STS pretraining.

\paragraph{Optimization and numerical settings.}
Each training stage processes 313,486 examples with gradient accumulation over four examples, yielding 78,372 optimizer updates; the final window contains two examples. AdamW uses frontend and Thinker learning rates of $2\times10^{-5}$ and $2\times10^{-6}$, weight decay $0.01$, and gradient clipping at norm one. Each stage uses 1\% warmup and cosine decay to 10\% of its initial learning rate. Parameters and the encoder remain FP32; frontend/Thinker computation uses BF16 autocast, with dropout disabled. Gaussian rewards clamp log inputs at $10^{-12}$ and norm denominators at $10^{-9}$.

\paragraph{Online and offline context.}
Online inference retains a 40-second audio history and trims only at completed boundaries. Offline evaluation uses the constructed complete prefixes under the 40,960-token context limit. These are different history policies; the offline results do not measure the effect of online history compression.

\paragraph{Evaluation consistency.}
For each endpoint, evaluation checks unique and complete test-ID coverage, agreement with the fixed targets and metadata, finite normalized action probabilities, and consistency between the predicted action and the probability argmax. Aggregate metrics are recomputed from saved predictions. These checks establish consistency of the reported evaluation; they do not independently validate the weak labels. The retained study artifacts include source-linked manifests, construction code, saved weak annotations, training configurations, and aggregate evaluations. Source permissions are heterogeneous: Wakeup and SmoothConv are recorded as CC BY-NC 4.0, Legacy turn text is an internal unpublished source whose underlying rights are not established in the current ledger, and FD v3 has no independent license record in the retained archive. Access to recordings and initialization weights therefore remains subject to their availability, applicable licenses, and internal authorization. The corpus revision is fixed on October 23, 2026; upstream source citations do not constitute a public release of the derived bilingual audio or action labels. A complete redistributable package is not provided with this manuscript, and the paper does not grant permission for commercial training, deployment, or redistribution.

\section{API Evaluation and Retries}\label{app:api-recovery}
The API evaluation uses a fixed set of 2,000 test IDs. Initial service failures affected 154 IDs. Retrying these IDs recovered 152 predictions, and two further requests recovered the remaining outputs, requiring 156 additional inference requests in total. No substitute examples were selected, and all 1,846 initially successful predictions were retained unchanged. The final evaluation contains 2,000 valid predictions with no unresolved service failures or invalid answers.

For the two remaining audio cases, the final audio chunk was merged into the preceding chunk while keeping each submitted chunk within the 30-second limit. Only transmission boundaries changed: the complete PCM input, sample order, model identifier, prompt, candidate order, and answer parser remained unchanged. The retry procedure therefore preserves the evaluated examples and successful original answers.

\section{Historical Text JEV: Supporting Diagnostic}\label{app:legacy-text}
The archived transcript-only JEV V7-B documents why a native three-action diagnostic cannot directly rank a five-action model. A decoder-only prompt supplies assistant text, the user segment, and action descriptions; fixed A/B/C logits represent \texttt{continue}, \texttt{yield}, and \texttt{wait}. No waveforms are supplied. The decoder and readout are jointly adapted, and the historical continuation retains supervision and masked reference preservation:
\begin{equation}
 \mathcal L_{\mathrm{text}}=\mathcal L_G+\mathcal L_{\mathrm{SFT}}
 +2m_iD_{\mathrm{KL}}(p_{\mathrm{V4}}\|p_\theta),
\end{equation}
Here $m_i$ enables reference preservation on 21,154 original rows and disables it on 8,000 augmented rows. Earlier initialization weights were removed, preventing complete replay of the parent chain.

On 1,606 matched Chinese legacy development examples, V7-B makes 1,580 correct decisions (98.38\% accuracy, 98.40\% macro-F1; Table~\ref{tab:text-native}). This is not a new blind test or an audio evaluation.

\begin{table}[H]
\centering\small
\caption{Historical Text JEV V7-B diagnostic under its native three-class targets. The set is development data, not the new Omni-JEV test set.}
\label{tab:text-native}
\begin{tabular}{lrrr}
\toprule
Model & $N$ & ACC & Macro-F1\\
\midrule
Text JEV V7-B & 1,606 & 98.38 & 98.40\\
\bottomrule
\end{tabular}

\end{table}

Yielding while idle differs from stopping active playback, and \texttt{wait} need not mean \texttt{listen}. Projecting the revised five-action targets to three labels disagrees with 618 original targets. This mismatch prevents direct ranking against Omni-JEV; a prospective standalone text test remains missing.

\paragraph{Illustration provenance.}
The architecture and RLCD schematics use AI-assisted artwork that was checked against the implemented computation. The sample blocks and probability bars are illustrative; numerical result plots are generated from the saved evaluation aggregates.

\end{document}